\documentclass[conference]{IEEEtran}
\IEEEoverridecommandlockouts

\usepackage{amsthm,amsmath}
\usepackage[utf8]{inputenc}
\usepackage{import}
\usepackage{graphicx}
\usepackage{rotating}
\usepackage[graphicx]{realboxes}
\graphicspath{ {images/} }
\usepackage{array}
\usepackage[citestyle=numeric-comp,bibstyle=ieee,%
backend=biber,sortcites]{biblatex}
\usepackage{orcidlink}
\usepackage{booktabs}

\usepackage{enumitem}
\setlist{nosep} 

\begin{document}

\title{Cross-Platform E-Commerce Product Categorization and Recategorization: A Multimodal Hierarchical Classification Approach\thanks{This work was funded by Fundação para a Ciência e a Tecnologia (UIDB/00124/2025,
UID/PRR/124/2025, Nova School of Business and Economics) and LISBOA2030
(DataLab2030 - LISBOA2030-FEDER-01314200).}}

\author{%
\IEEEauthorblockN{Lotte Gross\IEEEauthorrefmark{1},
Rebecca Walter\IEEEauthorrefmark{1},
Nicole Zoppi\IEEEauthorrefmark{1},
Adrien Justus\IEEEauthorrefmark{1},\\
Alessandro Gambetti\IEEEauthorrefmark{1}\orcidlink{0000-0003-3389-3784},
Qiwei Han\IEEEauthorrefmark{1}\orcidlink{0000-0002-6044-4530},
Maximilian Kaiser\IEEEauthorrefmark{2}\orcidlink{0009-0007-4329-161X}}
\IEEEauthorblockA{\IEEEauthorrefmark{1}Nova School of Business and Economics, Universidade NOVA de Lisboa, Carcavelos, Portugal\\
Email: \{51403, 49895, 53854, 53148, gambetti.alessandro, qiwei.han\}@novasbe.pt}
\IEEEauthorblockA{\IEEEauthorrefmark{2}Universität Hamburg / Grips Intelligence, Hamburg, Germany\\
Email: maximilian.kaiser@uni-hamburg.de}
}

\maketitle 

\begin{abstract}
This study addresses critical industrial challenges in e-commerce product categorization, namely platform heterogeneity and the structural limitations of existing taxonomies, by developing and deploying a multimodal hierarchical classification framework. Using a dataset of 271,700 products from 40 international fashion e-commerce platforms, we integrate textual features (RoBERTa), visual features (ViT), and joint vision–language representations (CLIP). We investigate fusion strategies, including early, late, and attention-based fusion within a hierarchical architecture enhanced by dynamic masking to ensure taxonomic consistency. Results show that CLIP embeddings combined via an MLP-based late-fusion strategy achieve the highest hierarchical F1 (98.59\%),  outperforming unimodal baselines. To address shallow or inconsistent categories, we further introduce a self-supervised “product recategorization” pipeline using SimCLR, UMAP, and cascade clustering, which discovered new, fine-grained categories (e.g., subtypes of ``Shoes'') with cluster purities above 86\%. Cross-platform experiments reveal a deployment-relevant trade-off: complex late-fusion methods maximize accuracy with diverse training data, while simpler early-fusion methods generalize more effectively to unseen platforms. Finally, we demonstrate the framework’s industrial scalability through deployment in EURWEB’s commercial transaction intelligence platform via a two-stage inference pipeline, combining a lightweight RoBERTa stage with a GPU-accelerated multimodal stage to balance cost and accuracy. 
\end{abstract}


\section{Introduction}
In the ever-evolving e-commerce landscape, effective product organization is essential for accurate listings, discoverability, operational efficiency, and marketing strategies such as SEO \cite{Zahavy2018}. From a customer perspective, clear taxonomies enable intuitive navigation and improve satisfaction and loyalty \cite{girard2002influence}. Large platforms rely on automated categorization pipelines, while smaller retailers often depend on manual processes that are time-consuming, error-prone, and costly to scale \cite{Bi2020,gholamian-etal-2024-llm}. As product volumes and diversity continue to expand, scalable automated categorization has become an industrial necessity rather than an optional feature.  

\begin{figure}[t]
\includegraphics[width=0.5\textwidth]{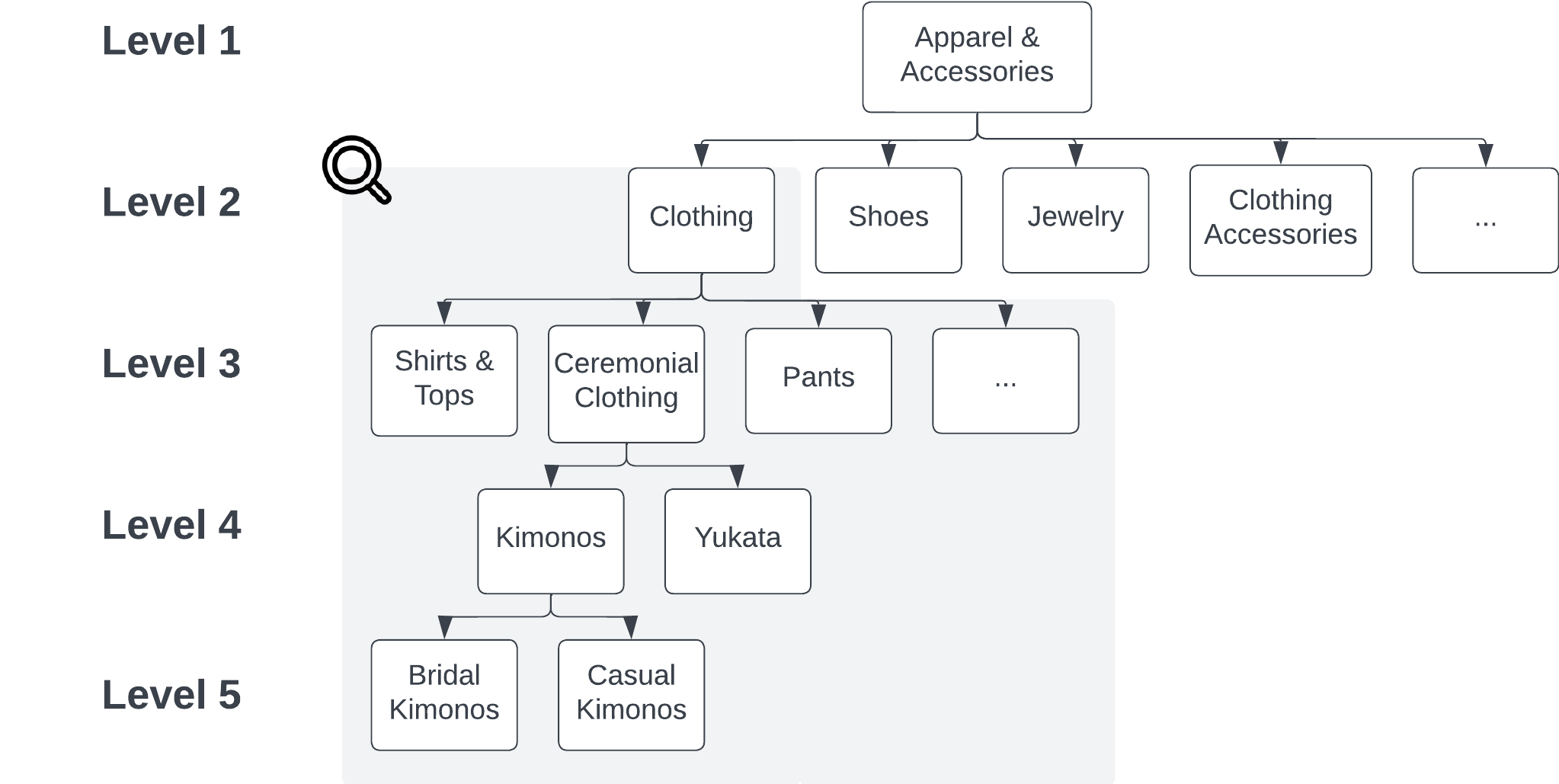}
\caption{Overview of the Google Product Taxonomy structure, showing imbalanced hierarchical levels for Clothing (5 levels) and Shoes (2 levels).} 
\label{fig:google_taxonomy}
\end{figure}

Despite its importance, automated categorization remains challenging. Smaller retailers often lack the resources and refined taxonomies of large platforms such as Amazon. Skewed product distributions hinder robust training for long-tail categories \cite{cheng2024commerce}, and inconsistent category standards across platforms \cite{avigdor2023consistent,zhang2021deep} complicate cross-platform analytics. Metadata itself is noisy, such as titles and descriptions vary widely in length, detail, and even language \cite{de2018generating}, making textual signals unreliable without complementary modalities. Proprietary taxonomies further exacerbate these issues, differing in depth and granularity (e.g., Google \cite{Krishnan2019}, Amazon \cite{Shen2011}). As shown in Figure \ref{fig:google_taxonomy}, the Google Product Taxonomy limits the \textit{Shoes} category to two levels, while other apparel categories extend to five. Such inconsistencies reduce the effectiveness of downstream applications: recommendation engines cannot distinguish between \textit{Sneakers} and \textit{Formal Shoes} if both are grouped under a shallow node. From an industrial perspective, this lack of granularity limits personalization, inventory analytics, and demand forecasting. Existing studies rarely address these platform-specific inconsistencies or the dynamic emergence of new categories \cite{chen-etal-2025-leveraging}, underscoring the need for adaptive, cross-platform solutions \cite{Cevahir2016}.  

To address these challenges, this paper presents a multimodal hierarchical product categorization framework designed for deployment at industrial scale, with a focus on the fashion e-commerce sector. Unlike purely academic studies, this work is motivated by collaboration with EURWEB, a commercial web intelligence provider, and has been integrated into their production pipeline. Our contributions are fourfold:

\begin{itemize}
    \item \textbf{Cross-Platform Data Utilization:} We leverage a dataset of 271,700 products spanning 40 e-commerce platforms, from global retailers like Zalando and Amazon to smaller platforms such as C\&A, Bonprix, and Peek\&Cloppenburg. This diversity improves robustness to heterogeneous metadata and enables smaller retailers to benefit from knowledge distilled from larger players.
    \item \textbf{Hierarchical Classification with Industrial Robustness:} We employ a hierarchical classification approach incorporating dynamic masking and shared hidden layers \cite{Ozyegen2022}. Rather than optimizing only for benchmark metrics, the design ensures taxonomic consistency and avoids invalid predictions—critical for downstream industrial use cases such as inventory management and recommendation systems.
    \item \textbf{Multimodal Architecture with Cost-Aware Fusion:} We integrate textual product attributes (titles and brands) with product images using RoBERTa, ViT, and CLIP. Multiple fusion strategies are evaluated to balance performance and inference cost \cite{Tashu2022}, supporting a two-stage inference pipeline: a lightweight RoBERTa model processes the majority of cases, while a multimodal model is selectively applied to low-confidence predictions, reducing infrastructure costs.
    \item \textbf{Self-Supervised Recategorization for Adaptive Taxonomies:} We introduce a scalable recategorization pipeline based on self-supervised learning and cascade clustering. This framework discovers new subcategories and refines existing ones (e.g., splitting \textit{Shoes} into \textit{Sneakers}, \textit{Boots}, and \textit{Open Shoes}), achieving stable purity levels above 85\%. For industry, this reduces reliance on manual relabeling and provides a semi-automated mechanism for adapting taxonomies to evolving product trends.
\end{itemize}

Through these contributions, our work goes beyond benchmark evaluation to demonstrate a deployable pipeline that balances accuracy, efficiency, and scalability. The EURWEB deployment validates the industrial feasibility of the approach, showing how multimodal hierarchical classification and taxonomy refinement can directly improve product discovery, recommendation quality, and sales forecasting. This study therefore contributes both to the research community and to practitioners seeking actionable insights into deploying Big Data solutions in industrial e-commerce settings.

\section{Related Work}

\subsection{Evolution of Product Categorization: From Unimodal to Multimodal Approaches}
Early approaches to product categorization were unimodal, relying on either text or images. Text-based methods analyzed titles and descriptions \cite{Cevahir2016,tan2020commerce}, with \cite{Kozareva2015} categorizing products using only short titles. Vision-based methods employed CNNs to detect main products in fashion images \cite{Yu2017}. While effective in narrow settings, text-only systems struggled with noisy or incomplete titles, whereas image-only methods were computationally costly for large catalogs. These limitations motivated multimodal approaches such as CDPF++ \cite{Kannan2011} and subsequent decision-level fusion \cite{Zahavy2018}, which showed that augmenting text with visual features outperformed unimodal baselines. Robustness was further improved by incorporating contextual text \cite{Kalva2007} and addressing labeling inconsistencies with semi-supervised augmentation \cite{avigdor2023consistent}.

Deep learning approaches, from BERT+ViT with cross-modal attention \cite{Chen2021} to pragmatic late-fusion strategies \cite{Tashu2022}, further advanced performance. Pre-trained vision–language models like CLIP \cite{radford2021learning} have been especially impactful: \cite{Fu2022} refined CLIP with cross-modality attention, and \cite{Gupta2016} applied Siamese networks with BERT and ResNet18 for product matching. Large datasets such as Rakuten \cite{Amoualian2021} enabled evaluation at scale, with \cite{Bi2020} confirming the industrial viability of late fusion.

Recently, Large Language Models (LLMs) have been explored for taxonomy tasks. \cite{cheng2024commerce} proposed a dual-expert framework combining domain-specific fine-tuning with general LLM reasoning. \cite{roumeliotis2025llms} compared GPT-4o and Claude 3.5 Sonnet for zero-shot categorization, while \cite{gholamian-etal-2024-llm} showed that in-context LLMs can surpass supervised models under noisy metadata. \cite{chen-etal-2025-leveraging} introduced an LLM-agnostic multimodal framework designed to enforce taxonomy consistency. However, LLM adoption in production is limited by inference cost, latency, privacy, and weak consistency guarantees. Supervised multimodal systems remain attractive due to their controllable inference footprint and compatibility with industrial infrastructure.

\subsection{Flat versus Hierarchical Product Categorization}
E-commerce catalogs typically employ hierarchical structures (e.g., Google Product Taxonomy \cite{Krishnan2019}), leading to two main strategies: flat classifiers, which assign items directly to leaf nodes \cite{cotacallapa2024flat}, and hierarchical classifiers, which traverse taxonomies level by level \cite{Silla2010}. Flat classifiers sometimes outperform but ignore parent–child consistency \cite{Krishnan2019}, while hierarchical methods enforce taxonomic coherence \cite{Bergamaschi2002,Hasson2021}.

Research on improving hierarchical classification spans multiple directions. \cite{Shen2011} improved eBay’s taxonomy accuracy with domain-specific features, and \cite{Gupta2016} used distributional semantics in a two-level ensemble. Deep learning introduced autoencoders for multi-level classification \cite{Cevahir2016}, hierarchical BERT-based models \cite{zhang2021deep}, and loss functions penalizing taxonomic errors \cite{Gao2020}. Practical deployments include CatReComm \cite{Hasson2021}, an interactive category recommendation system. Dynamic masking \cite{Ozyegen2022} further strengthened consistency by constraining fine-level predictions to valid children of predicted parent nodes.

Despite these advances, most existing approaches emphasize algorithmic performance over deployment concerns. For industry, scalability across heterogeneous taxonomies, robustness to noisy inputs, and inference efficiency are equally important. This gap motivates our work, which integrates multimodal learning, dynamic masking, and self-supervised recategorization into a deployable architecture addressing real-world constraints such as taxonomy heterogeneity, cost-efficiency, and evolving product categories.

\section{Data}
\label{sec:data}
The dataset for this study was obtained from a leading European Web intelligence provider (hereafter referred to as EURWEB). It consists of 271,700 fashion products belonging to the Level 1 category, ``Apparel \& Accessories,'' scraped from 40 prominent e-commerce platforms. These platforms span both large-scale retailers such as Zalando and Amazon, as well as smaller specialized retailers such as C\&A, Bonprix, and Peek\&Cloppenburg. The inclusion of both major and niche retailers was intentional: large platforms provide deep and diverse product taxonomies, while smaller platforms often contain inconsistent or incomplete product metadata. Together, they capture the heterogeneity and imbalance characteristic of real-world e-commerce data and reflect the deployment challenges faced by industrial systems.


Each product record is associated with metadata variables including URL, GTIN, Title, Brand, Currency, Price, SKU, Offer ID, Batch ID, and Image URL. In addition, all items were enriched with product categories based on the Google Product Taxonomy, a widely used hierarchical classification system in the retail industry. For example, one product might be assigned to the category \textit{Apparel \& Accessories $>$ Clothing $>$ Traditional \& Ceremonial Clothing $>$ Kimonos $>$ Bridal Kimonos}, spanning five levels of depth. Such hierarchical labeling provides a structured foundation for classification tasks, but also reveals systemic imbalances and limitations in standard taxonomies (e.g., shallow versus deep category trees).

To prepare the dataset for modeling, several preprocessing steps were undertaken with industrial deployment constraints in mind. First, product images were downloaded when available, and items with unidentifiable or corrupted images were discarded, ensuring that visual features would not degrade system robustness. Observations with missing values in key textual fields (title, brand, category) were removed to maintain data quality, reflecting the need for reliable inputs in production pipelines. Features not directly relevant to the categorization task (e.g., GTIN, Currency, SKU) were excluded to minimize noise and reduce computational overhead. Finally, the hierarchical category strings were decomposed into separate columns representing each level (``cat\_level\_1,'' ``cat\_level\_2,'' etc.), which facilitated structured training and prediction across multiple taxonomy depths. This transformation ensured compatibility with hierarchical models and simplified downstream system integration.



\subsection{Explorative Data Analysis} \label{sec:dataEDA}

As all products in the dataset belong to the Level 1 category ``Apparel \& Accessories,'' our analysis focuses on deeper levels of the taxonomy. The average hierarchy depth is Level 3, with 67.3\% of products categorized up to this level, but fewer than 7\% extend to Level 5. At Level 2, the taxonomy contains 8 broad categories with uneven expansion. For example, ``Shoes'' is consistently restricted to two levels, while ``Clothing'' extends to five levels, reflecting greater variety and granularity. Such inconsistencies highlight the need for adaptive recategorization methods capable of refining underdeveloped branches.

A second challenge is severe class imbalance. ``Shoes'' and ``Clothing'' together account for 84.28\% of all products, while categories such as ``Shoe Accessories'' represent less than 0.1\%. This imbalance propagates into deeper levels, where ``Clothing'' dominates with 17 of the 56 Level 3 subcategories (e.g., ``Shirts \& Tops,'' ``Pants,'' ``Outerwear''), while ``Shoes'' remains flat. From an industrial perspective, this means models must be robust to underrepresented categories while still preserving accuracy for common ones.

Two additional sources of heterogeneity are textual and visual attributes. Product titles average 37 characters but range from 2 to 255, with strong lexical skew (e.g., frequent gender-specific tokens and clothing descriptors). Brand distributions are similarly imbalanced: TeePublic lists only apparel, while Nike balances apparel and shoes, reinforcing categorical skew. Image quality was also evaluated using background-cleanliness measures. Across platforms, 97\% of images had clean backgrounds, supporting reliable vision pipelines. However, some retailers deviated (e.g., Fashion Nova) had 84\% unclean image, while categories such as ``Lingerie'' and ``Wristbands'' also showed noisier presentation styles). For industrial deployment, this indicates that while most imagery is standardized, robust preprocessing is needed for noisy subsets.

\section{Methodology}
\label{sec:multimodalLearning}

\begin{figure*}[t]
\centering
\includegraphics[width=0.65\textwidth]{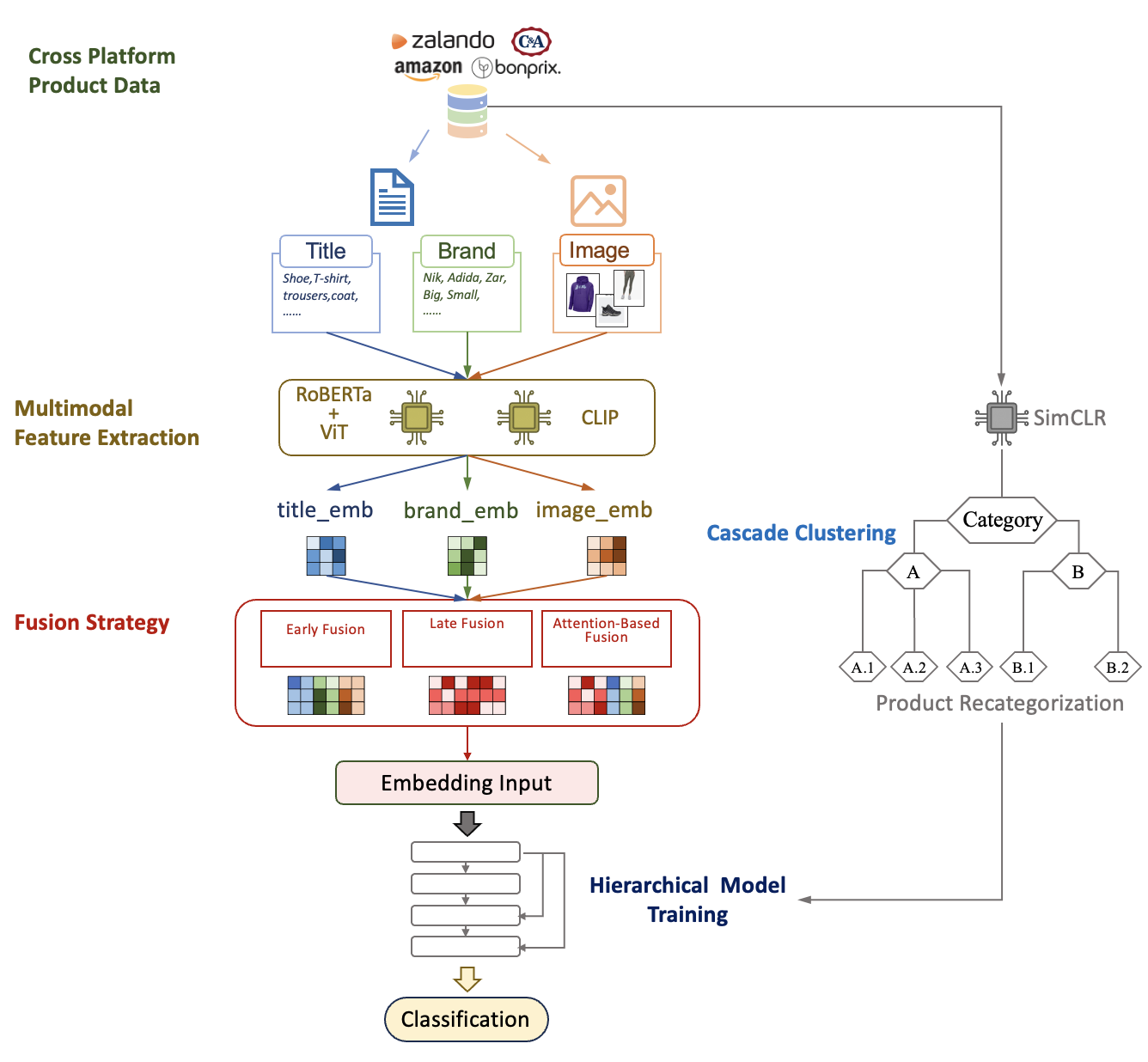}
\caption{Overview of the proposed methodology, illustrating the pipeline from cross-platform data input through multimodal feature extraction (RoBERTa, ViT, CLIP), multiple fusion strategies, and hierarchical classification, alongside the parallel process for product recategorization using SimCLR and cascade clustering.} \label{fig:pipeline}
\end{figure*}

Multimodal product categorization in this study involves three main stages: (1) extraction and processing of features from textual and visual data using modality-specific and multimodal techniques; (2) learning multimodal representations, where signals from different modalities are fused and aligned into a common space; and (3) hierarchical multi-label classification, where the fused features are used to predict product categories across multiple taxonomy levels. An overview of this pipeline is illustrated in Figure \ref{fig:pipeline}.  

While initial model optimization and feature extractor selection were performed on a subset of data from the Zalando platform (chosen due to its consistency, as described in Section \ref{sec:dataEDA}), all final experiments were conducted both on Zalando and on the complete cross-platform dataset of 40 retailers. This dual evaluation allowed us to measure not only peak classification accuracy but also robustness and generalization across heterogeneous platforms, which is an essential requirement for production deployment in industry.

\subsection{Feature Extraction}
\label{sec:embeddingAlgorithms}

Feature extraction was a critical design choice, as industrial deployment requires models that balance representational power with inference efficiency. Based on preliminary experiments, we selected RoBERTa for text embeddings and ViT for image embeddings because of their superior accuracy and stability in flat classification compared to other candidates. Table \ref{tab:flat_merged_results} summarizes the comparative performance, showing that RoBERTa consistently outperformed BERT, ALBERT, and DistilBERT on textual data, while ViT surpassed CNN-based models such as VGG16 and ResNet50 on images.

\begin{table*}[ht]
\centering
\caption{Summary of results for flat classification. The best model for each modality was selected as the feature extractor in the multimodal pipeline.}
\label{tab:flat_merged_results}
\begin{tabular}{l|cccc|ccc}
\toprule
 & \textbf{BERT} & \textbf{ALBERT} & \textbf{DistilBERT} & \textbf{RoBERTa} & \textbf{VGG16} & \textbf{ResNet50} & \textbf{ViT} \\
\midrule
Accuracy  & 92.89\% & 94.28\% & 93.60\% & \textbf{95.10\%} & 84.10\% & 84.57\% & \textbf{86.69\%} \\
Precision & 92.29\% & 92.42\% & 93.09\% & \textbf{94.78\%} & 83.27\% & 83.84\% & \textbf{85.90\%} \\
Recall    & 92.89\% & 94.28\% & 93.60\% & \textbf{95.10\%} & 84.10\% & 84.57\% & \textbf{86.69\%} \\
Weighted F1 & 92.37\% & 93.27\% & 93.22\% & \textbf{94.66\%} & 83.12\% & 83.86\% & \textbf{86.09\%} \\
\bottomrule
\end{tabular}
\end{table*}

For textual attributes, both product titles and brands were encoded separately using a 125M parameter pre-trained RoBERTa model \cite{liu2019roberta}. Embeddings were obtained by extracting the last hidden state and computing the mean across tokens, yielding a dense vector representation for each attribute. This design reflects real-world considerations, as brand and title often provide complementary but noisy signals in industrial datasets. Image features were extracted with a ViT-B/16 model \cite{dosovitskiy2021image}, chosen for its ability to capture fine-grained fashion attributes such as patterns and cuts. To reduce inference cost, Principal Component Analysis (PCA) was applied to the image embeddings, preserving 90\% of the variance while reducing dimensionality. This trade-off between representational fidelity and computational efficiency proved critical for deployment in GPU-constrained environments.

As an alternative, we also employed CLIP (ViT-B/32) \cite{radford2021learning}, a pre-trained vision-language model that aligns text and images in a shared embedding space. CLIP’s parallel encoders were used to separately process titles, brands, and images. Normalized embeddings were extracted for each modality. Unlike RoBERTa+ViT, CLIP performs alignment during pre-training, making it “fusion-free” in principle. However, downstream classification still requires combining these embeddings, so additional fusion mechanisms were evaluated. The use of CLIP highlights a trade-off familiar to practitioners: pretrained multimodal alignment offers strong generalization but can be more costly to fine-tune at industrial scale.

\subsection{Fusion Strategies}
\label{sec:multimodalLearningFusionStrategies}

After extracting unimodal embeddings (RoBERTa for text, ViT for images) or aligned CLIP embeddings, the next design decision was how to fuse them into a joint representation suitable for hierarchical classification. We systematically explored multiple strategies, reflecting a spectrum from computationally lightweight to more expressive but costlier mechanisms:

\begin{itemize}
\item \textbf{Simple Concatenation (Early Fusion):} Title, brand, and image embeddings are concatenated into a single vector. This approach is computationally efficient, making it attractive for industry deployments requiring low-latency inference.
\item \textbf{Feature Transformation with MLP (Late Fusion):} Each modality is passed through a separate MLP with ReLU activation before concatenation. The transformed embeddings are optionally processed by an additional MLP prior to classification. This design allows the system to learn modality-specific transformations, balancing accuracy gains with modest increases in inference cost.
\item \textbf{Cross-Attention Fusion (Attention-Based Fusion):} To explicitly model inter-modality dependencies, we implemented cross-attention mechanisms between pairs of embeddings (e.g., brand–image, title–image). The outputs were concatenated with the original embeddings and processed by MLPs. While more resource-intensive, this approach captures fine-grained interactions critical for ambiguous cases (e.g., differentiating ``Running Shoes'' from ``Sneakers''). 
\end{itemize}


\subsection{Hierarchical Model Architecture}
\label{sec:multimodalHierarchicalModelArchitecture}

To classify products into the multi-level taxonomy, we developed a hierarchical neural network architecture that integrates multimodal embeddings with structured prediction. Unlike flat classifiers, which assign items directly to leaf nodes, this architecture explicitly traverses the taxonomy, ensuring predictions remain consistent with the hierarchy. Such consistency is crucial in industrial contexts, where invalid category paths (e.g., predicting \textit{Shoes $\rightarrow$ Dresses}) are unacceptable for downstream search and recommendation engines.

\begin{figure*}[t]
\centering
\includegraphics[width=0.25\linewidth, angle=90]{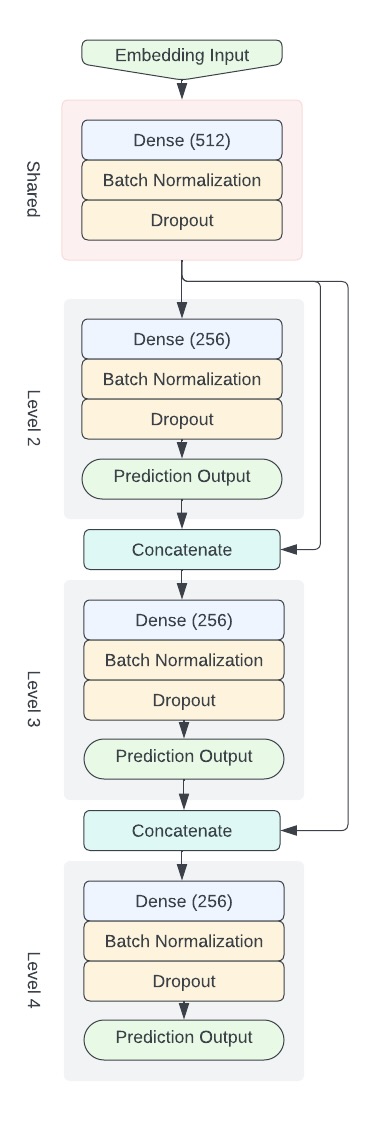}
\caption{Hierarchical model architecture. A shared layer processes the multimodal embedding. Subsequent layers combine shared representations with coarser predictions before making finer predictions.} \label{fig:hierarchical_model}
\end{figure*}

The model begins with a shared Dense layer (512 neurons, ReLU activation), followed by Batch Normalization and Dropout for regularization. The first level of the taxonomy is predicted using this shared representation, passed through an additional Dense layer (256 neurons, ReLU, Batch Normalization, Dropout) and a Softmax classifier. For subsequent levels, predictions from the coarser level are concatenated with the shared features and passed through similar blocks, enabling finer-level predictions to leverage both general features and hierarchical context.

\begin{figure}[t]
\centering
\includegraphics[width=\linewidth]{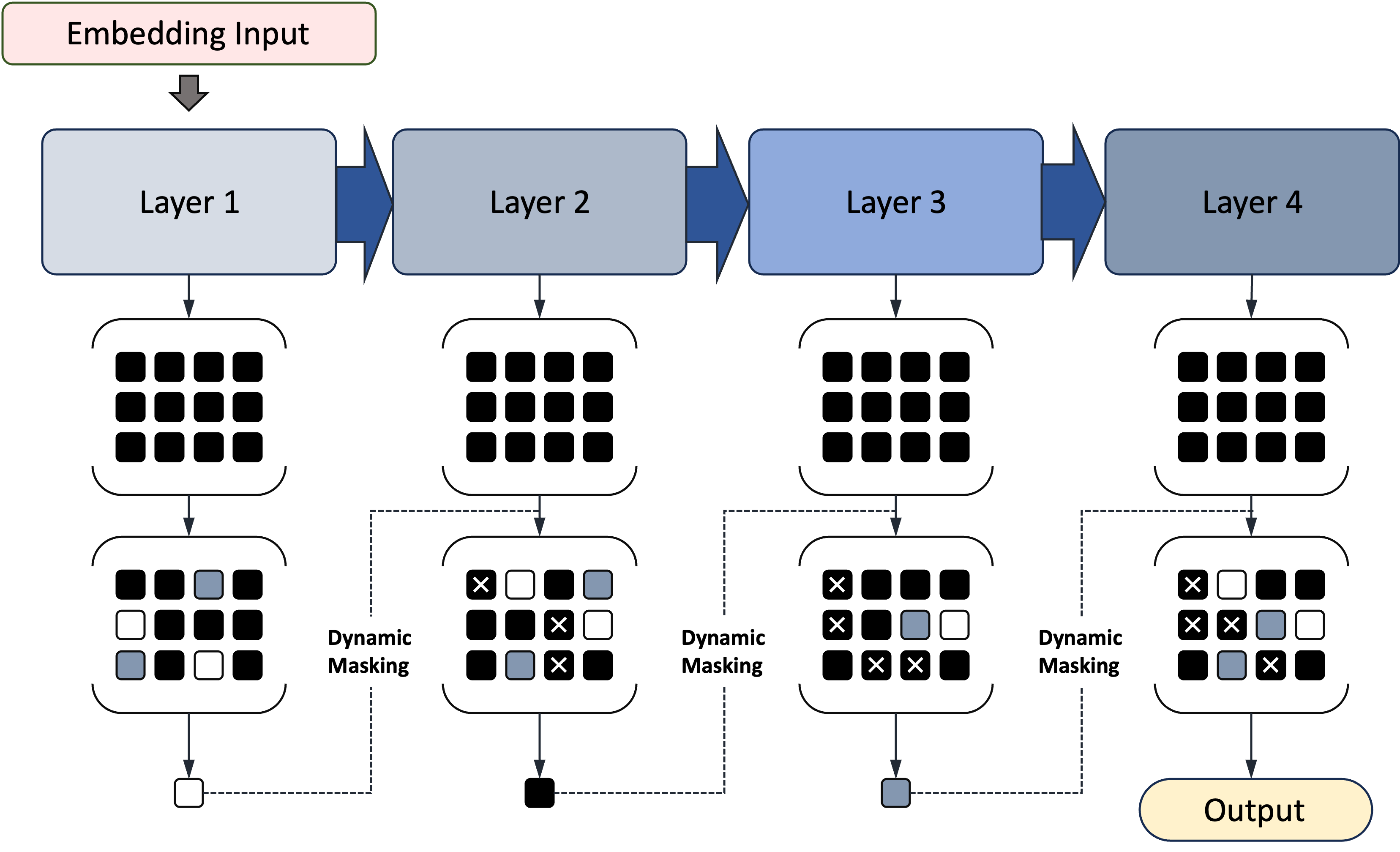}
\caption{Dynamic masking mechanism applied between hierarchical prediction layers. Each coarser-level prediction constrains the candidate set at the next level, ensuring valid taxonomy paths.} \label{fig:dynamic_masking}
\end{figure}

A central innovation in this architecture is \textbf{Dynamic Masking}, illustrated in Figure \ref{fig:dynamic_masking}. Following \cite{Ozyegen2022}, dynamic masking constrains predictions at finer levels by filtering out categories inconsistent with the predicted parent. In practice, once a parent category is selected, a binary mask zeros out non-child categories in the Softmax layer of the next level. This ensures all predicted paths are taxonomically valid and reduces classification noise. From an industrial standpoint, dynamic masking enforces structural consistency without requiring costly post-processing, thereby improving both accuracy and interpretability.

For optimization, models were trained with the Adam optimizer (initial learning rate 1e-3) and categorical cross-entropy loss, using stratified sampling to split the dataset into training (64\%), validation (16\%), and test (20\%). Early stopping with a patience of 5 epochs was used to avoid overfitting. In practice, models converged within 60 epochs with a batch size of 128. These settings reflect a balance between accuracy and computational cost, tuned to be viable in an industrial deployment pipeline where retraining on new product batches is expected.

\subsection{Product Recategorization}
\label{sec:multimodalLearningSSL}

Standard product taxonomies such as the Google Product Taxonomy provide a useful baseline but often exhibit structural limitations, particularly in depth and adaptability. Categories such as ``Shoes'' typically lack the granularity present in other apparel categories (e.g., ``Clothing''), making it difficult for platforms to capture fine-grained distinctions important for search, personalization, and inventory management.  

To address these shortcomings, we developed a \textit{product recategorization} methodology designed to refine existing taxonomies and systematically identify new subcategories. The approach balances two competing requirements: (1) sufficient granularity to capture meaningful distinctions for end users and businesses, and (2) cross-platform generalizability so that refinements discovered on one dataset (e.g., Zalando) can be applied consistently across heterogeneous retailers. The framework integrates self-supervised learning (SSL) for robust feature representation with unsupervised clustering for subcategory discovery, resulting in a scalable and semi-automated taxonomy refinement pipeline.

The methodological workflow consists of the following steps:

\begin{enumerate}
\item \textbf{Image Filtering for Clustering Purity:} The process begins by identifying visually consistent images suitable for clustering. Image embeddings are generated with SimCLR, a self-supervised model trained to capture fine-grained visual similarity. These embeddings are clustered (e.g., with K-Means) to separate visually coherent products (e.g., similar angles, standardized backgrounds) from noisy cases. This filtering step is critical for ensuring that subsequent cluster formation is meaningful, reflecting industrial requirements for stable, high-purity categories.
\item \textbf{Dimensionality Reduction:} High-dimensional embeddings from text and images are computationally expensive for clustering. To manage scalability, we applied UMAP (Uniform Manifold Approximation and Projection), which preserves both local and global neighborhood structures in the embedding space. UMAP’s ability to generate low-dimensional yet structure-preserving embeddings allows clusters to reflect true category distinctions without overwhelming infrastructure resources.
\item \textbf{Cascade Clustering for Taxonomy Refinement:} To iteratively refine taxonomies, we introduced a cascade clustering procedure. First, UMAP is applied to embeddings of a target product set (e.g., all ``Shoes''). Agglomerative hierarchical clustering is then used to identify subgroups within this reduced embedding space. These clusters are visualized and validated using domain knowledge and sample inspection, enabling the assignment of meaningful subcategory labels. Through this iterative process, general categories (e.g., ``Shoes'') are subdivided into fine-grained categories (e.g., ``Sneakers,'' ``Boots,'' ``Open Shoes''), addressing the imbalance and shallowness of existing taxonomies.
\item \textbf{Generalizing New Taxonomies Across Platforms:} Once discovered, the new subcategories are used to retrain the hierarchical classification model (Section \ref{sec:multimodalHierarchicalModelArchitecture}). This extended classifier can then generalize the refined taxonomy across all platforms, enabling scalable deployment without requiring re-clustering for each platform individually. This generalization step ensures that recategorization benefits are transferable across diverse datasets, a key requirement for industrial adoption.
\end{enumerate}

This methodology creates a bridge between unsupervised discovery and supervised classification, enabling taxonomies to evolve semi-automatically. From an industry perspective, it provides a scalable alternative to manual taxonomy engineering, reducing dependence on costly human annotation and enabling rapid adaptation to emerging product types.

\begin{figure}[t]
\centering
\includegraphics[width=\linewidth]{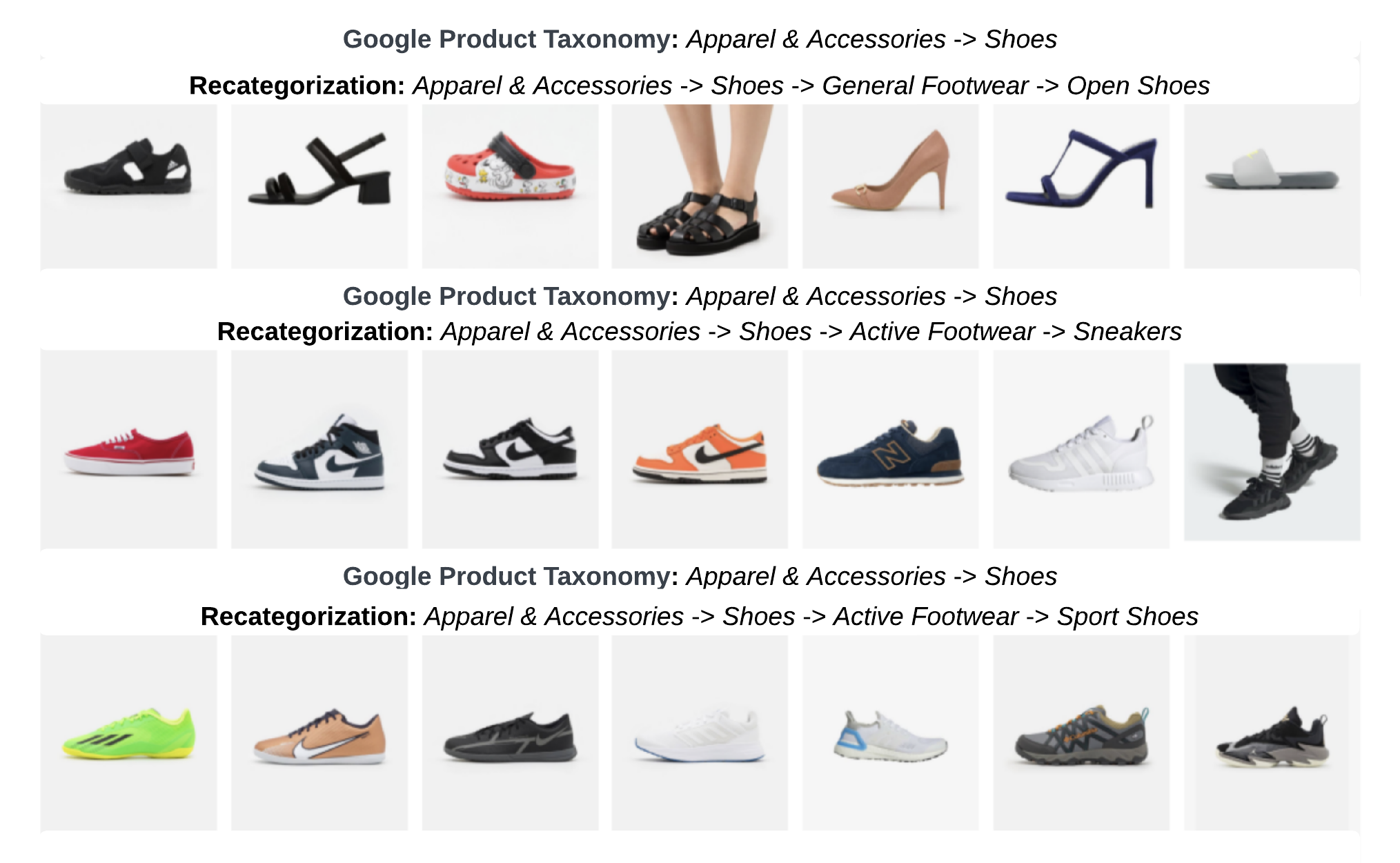}
\caption{Illustrative example of recategorization for ``Open Shoes,'' ``Sneakers,'' and ``Sport Shoes'' derived from the broader ``Shoes'' category. This process demonstrates how the pipeline refines shallow taxonomies into more granular, industrially useful categories.} \label{fig:recategorization}
\end{figure}

\subsection{Evaluation Metrics}
\label{sec:multimodalLearningEvaluationMetrics}

To evaluate model performance, particularly for hierarchical classification tasks, we adopted a combination of traditional and hierarchy-aware metrics. Standard flat classification measures such as F1-score were used in both weighted and macro variants: the weighted F1 accounts for class imbalance, while the macro F1 provides equal importance to all categories, regardless of frequency. These metrics provide baseline interpretability for practitioners familiar with conventional classification evaluations.  

However, standard flat metrics alone are insufficient to evaluate hierarchical models, as they do not account for partial correctness in taxonomy-structured predictions \cite{Silla2010}. For example, predicting ``Shoes'' instead of ``Sneakers'' should not be penalized as harshly as predicting ``Handbags.'' To capture this nuance, we incorporated hierarchy-aware metrics: hierarchical precision (hP), hierarchical recall (hR), and hierarchical F-score (hF) \cite{miranda2023hiclass}. These metrics explicitly consider taxonomy paths when computing overlaps between predictions and ground truth, aligning better with real-world expectations for classification consistency in industrial settings.

Formally, the metrics are defined as follows:
\begin{equation} \label{eq:hP}
hP = \frac{\sum_i | \alpha_i \cap \beta_i |}{\sum_i |\alpha_i|}
\end{equation}
\begin{equation} \label{eq:hR}
hR = \frac{\sum_i | \alpha_i \cap \beta_i |}{\sum_i |\beta_i|}
\end{equation}
\begin{equation}
hF = \frac{2 \times hP \times hR}{hP + hR}
\end{equation}
where, for each test example $i$, $\alpha_i$ is the set of predicted classes at the most specific level and all their ancestor classes in the taxonomy, and $\beta_i$ is the corresponding set of true classes and their ancestors. Summations are taken over all test examples.  

By employing both flat and hierarchical metrics, we ensured that evaluation captured not only overall classification accuracy but also the structural validity of predictions—an aspect critical for industrial deployment, where invalid paths can propagate errors into downstream applications such as search indexing, inventory analytics, and personalized recommendations.

\subsection{Unimodal Benchmark Models}
\label{sec:resultsBenchmark}

To contextualize the performance of our multimodal architecture, we implemented several unimodal benchmark models. These baselines serve two purposes: (1) they provide a point of comparison for evaluating the contribution of multimodality, and (2) they reflect simplified versions of models that smaller retailers could feasibly deploy in resource-constrained environments.  

\begin{enumerate}
\item \textbf{Text-Only:} A model that relies exclusively on textual features (titles and brands). Text embeddings were generated with RoBERTa and fed into the hierarchical classification head. This benchmark reflects the most common industrial baseline, as many retailers have textual but not visual data pipelines in place.
\item \textbf{Image-Only:} A model using only visual features extracted by ViT (with dimensionality reduced via PCA). This benchmark isolates the visual modality and is relevant for scenarios where textual metadata is sparse, incomplete, or unreliable—frequent issues in cross-platform deployments.
\item \textbf{Self-Supervised Image Model:} A benchmark using SimCLR with a ResNet-18 backbone (implemented with the Lightly SSL framework). A projection head maps embeddings into a 128-dimensional latent space optimized with NTXentLoss. Multiple augmentation intensities (``Low,'' ``Middle,'' ``High'') were tested to evaluate robustness to presentation noise. While more resource-intensive, this benchmark demonstrates the potential of SSL approaches in contexts where annotated product categories are scarce.
\end{enumerate}

Together, these benchmarks provide a spectrum of unimodal strategies, ranging from lightweight text-only pipelines to more resource-intensive SSL image models. Comparing them to our multimodal architecture highlights not only accuracy improvements but also the trade-offs that industrial practitioners must consider when designing categorization systems under real-world constraints such as annotation cost, computational budget, and deployment scalability.

\section{Results}
\label{sec:results}

\subsection{Performance of Fusion Strategies}

Table \ref{tab:comprehensive_performance} summarizes the performance of unimodal baselines and multimodal models with different fusion strategies. For CLIP embeddings, the ``Late-Fusion'' approach (MLP-based feature transformation) consistently achieved the strongest results, reaching an hF1 of 98.59\% and maintaining very high Macro F1 across all taxonomy depths (e.g., 99.32\% at Level 2, 98.31\% at Level 3, and 98.23\% at Level 4). These scores illustrate the ability of CLIP-based multimodal embeddings to handle class imbalance and achieve consistent generalization across multiple hierarchical levels. In contrast, ``Early-Fusion'' with CLIP (concatenation) also performed well (hF1 97.70\%), but with slightly lower robustness at finer levels.  

For the RoBERTa \& ViT combination, ``Late-Fusion'' again provided the best performance (hF1 95.54\%), outperforming both ``Early-Fusion'' (hF1 90.85\%) and ``Attention-Based Fusion'' (hF1 90.66\%). Interestingly, the attention-based fusion mechanisms underperformed despite their higher complexity, suggesting that more expressive architectures do not necessarily translate into practical gains for heterogeneous, industry-scale data. This finding is particularly relevant in industrial contexts where additional computational cost must be justified by measurable performance improvements.  

Overall, across all assessed strategies, multimodal approaches using CLIP embeddings clearly outperformed those relying on separately trained RoBERTa and ViT models. The CLIP ``Late-Fusion'' configuration emerged as the most effective, striking a balance between industrial robustness, scalability, and classification accuracy.

\begin{table*}[t]
\centering
\caption{Summary of hierarchical classification results for unimodal benchmarks and multimodal approaches with different fusion strategies. "Early-Fusion" corresponds to Simple Concatenation, "Late-Fusion" to Feature Transformation with MLP, and "Attention-Based Fusion" to Feature Transformation with Cross-Attention. Best overall multimodal performance highlighted in bold.}
\label{tab:comprehensive_performance}
\resizebox{\textwidth}{!}{
\begin{tabular}{lccccccccc}
\toprule
\textbf{Metric} & \textbf{RoBERTa} & \textbf{ViT} & \textbf{SimCLR} & \multicolumn{3}{c}{\textbf{RoBERTa \& ViT}} & \multicolumn{3}{c}{\textbf{CLIP}} \\
\cmidrule(lr){5-7} \cmidrule(lr){8-10}
 & & & & \textbf{\begin{tabular}[c]{@{}c@{}}Early-\\Fusion\end{tabular}} & \textbf{\begin{tabular}[c]{@{}c@{}}Late-\\Fusion\end{tabular}} & \textbf{\begin{tabular}[c]{@{}c@{}}Attention-Based\\Fusion\end{tabular}} & \textbf{\begin{tabular}[c]{@{}c@{}}Early-\\Fusion\end{tabular}} & \textbf{\begin{tabular}[c]{@{}c@{}}Late-\\Fusion\end{tabular}} & \textbf{\begin{tabular}[c]{@{}c@{}}Attention-Based\\Fusion\end{tabular}} \\
\midrule
\textbf{Macro F1} & & & & & & & & & \\
Level 2 & 82.92\% & 75.23\% & 62.92\% & 77.33\% & 85.38\% & 70.37\% & 99.00\% & \textbf{99.32\%} & 98.66\% \\
Level 3 & 61.03\% & 50.43\% & 30.79\% & 50.67\% & 65.56\% & 43.74\% & 97.15\% & \textbf{98.31\%} & 95.28\% \\
Level 4 & 54.27\% & 42.65\% & 23.87\% & 42.12\% & 57.15\% & 36.20\% & 97.03\% & \textbf{98.23\%} & 95.06\% \\
\textbf{Weighted F1} & & & & & & & & & \\
Level 2 & 95.75\% & 96.71\% & 93.35\% & 96.66\% & \textbf{98.22\%} & 96.32\% & 92.55\% & 95.22\% & 86.85\% \\
Level 3 & 90.74\% & 87.77\% & 77.74\% & 87.72\% & \textbf{94.37\%} & 87.95\% & 82.15\% & 88.49\% & 67.21\% \\
Level 4 & 90.45\% & 87.39\% & 77.11\% & 87.28\% & \textbf{94.04\%} & 87.35\% & 75.64\% & 82.28\% & 57.68\% \\
\textbf{hP} & 92.39\% & 91.23\% & 83.82\% & 91.11\% & 95.68\% & 91.08\% & 97.78\% & \textbf{98.66\%} & 96.47\% \\
\textbf{hR} & 92.44\% & 90.62\% & 83.71\% & 90.59\% & 95.39\% & 90.24\% & 97.62\% & \textbf{98.52\%} & 96.25\% \\
\textbf{hF1} & 92.41\% & 90.92\% & 83.77\% & 90.85\% & 95.54\% & 90.66\% & 97.70\% & \textbf{98.59\%} & 96.36\% \\
\bottomrule
\end{tabular}
}
\end{table*}

\subsection{Comparison with Unimodal Benchmarks}

The superiority of the multimodal architecture becomes evident when compared to unimodal benchmarks. The best multimodal model (CLIP + ``Late-Fusion'') achieved an hF1 of 98.59\%, substantially higher than the RoBERTa-only model (92.41\%), the ViT-only model (90.92\%), and the SimCLR self-supervised image model (83.77\%). These results highlight the value of combining modalities in industrial contexts where product metadata is often noisy, incomplete, or inconsistent.  

In terms of Macro F1, CLIP ``Late-Fusion'' achieved 99.32\% at Level 2, 98.31\% at Level 3, and 98.23\% at Level 4, consistently outperforming unimodal alternatives. For Weighted F1, the RoBERTa \& ViT ``Late-Fusion'' setup yielded competitive results (98.22\% at Level 2, 94.37\% at Level 3, 94.04\% at Level 4), suggesting that while multimodal embeddings improve balanced performance across classes, while supervised unimodal models can still perform strongly for frequent categories. This distinction is important for deployment: organizations with limited infrastructure may initially adopt unimodal systems, but multimodal late-fusion architectures deliver superior performance when accuracy across long-tail categories is critical.

\subsection{Product Recategorization Results}

The product recategorization pipeline (Section \ref{sec:multimodalLearningSSL}) was applied to the ``Shoes'' category, which in the Google Product Taxonomy is underdeveloped compared to ``Clothing.'' Using ResNet-50 and BERT embeddings with cascade clustering, we identified seven new subcategories spanning two additional hierarchical levels, including ``Sneakers,'' ``Sport Shoes,'' ``Boots,'' and ``Open Shoes.''  

Evaluation on a filtered Zalando dataset showed promising results. An illustrative example of how this method generates new shoe subcategories is shown in Figure \ref{fig:recategorization}. At Level 3, the new clusters (e.g., ``Active Footwear,'' ``General Footwear'') achieved an average purity of 89\%. At Level 4, more fine-grained subcategories (e.g., ``Sneakers,'' ``Boots'') maintained purity at 86.4\%. When the hierarchical classifier was retrained with these new labels and tested across the full Zalando shoe dataset (including noisier images), the purity remained stable at ~85\%. This demonstrates that the recategorization framework not only discovers meaningful and domain-relevant categories but also produces subcategories robust enough to generalize across platforms. From an industry standpoint, this reduces reliance on manual taxonomy engineering and provides a scalable solution to maintaining up-to-date product hierarchies.

\subsection{Cross-Platform Generalization of Models}
\label{sec:resultsReclassification}

A key objective of this study was to test whether models trained on a single platform can generalize to unseen platforms—a realistic industrial challenge where retraining across all sources may not always be feasible. Table \ref{tab:cross_platform_results} presents cross-platform inference results for models trained solely on Zalando and tested on 39 other platforms.  

Among the evaluated models, CLIP with ``Early-Fusion'' achieved the highest generalization (hF1 93.78\%), outperforming both CLIP ``Late-Fusion'' (91.88\%) and the unimodal ViT (86.63\%). While ``Late-Fusion'' yielded superior accuracy when trained on diverse multi-platform data, its performance dropped more sharply in the cross-platform setting. By contrast, ``Early-Fusion'' exhibited greater robustness when exposed to unseen domains.  

These findings highlight a trade-off directly relevant to deployment. Complex fusion strategies excel when trained on large and diverse datasets, but simpler strategies may generalize better under data-limited conditions. In practice, organizations deploying models in new markets or smaller platforms without extensive labeled data may benefit from lightweight fusion methods, while large retailers with access to diverse data can maximize accuracy using more advanced late-fusion approaches.

\begin{table}[t]
\centering
\caption{Cross-platform inference results: models trained on Zalando data and tested on other platforms.}
\label{tab:cross_platform_results}
\begin{tabular}{lccc}
\toprule
\textbf{Metric} & \textbf{ViT} & \textbf{CLIP Concat} & \textbf{CLIP MLP} \\
\midrule
\textbf{Macro F1} & & & \\
Level 2 & 82.87\% & \textbf{91.70\%} & 84.19\% \\
Level 3 & 51.73\% & \textbf{74.57\%} & 72.06\% \\
Level 4 & 40.42\% & \textbf{61.79\%} & 61.22\% \\
\textbf{Weighted F1} & & & \\
Level 2 & 94.31\% & \textbf{97.04\%} & 95.59\% \\
Level 3 & 82.74\% & \textbf{92.48\%} & 92.17\% \\
Level 4 & 82.59\% & \textbf{92.24\%} & 91.71\% \\
\textbf{hP} & 87.08\% & \textbf{94.48\%} & 93.89\% \\
\textbf{hR} & 86.18\% & \textbf{93.08\%} & 90.01\% \\
\textbf{hF1} & 86.63\% & \textbf{93.78\%} & 91.88\% \\
\bottomrule
\end{tabular}
\end{table}

\section{Deployment}

The product categorization models and strategies developed in this study have been implemented in practice within EURWEB's international e-commerce transaction intelligence platform. This commercial platform aggregates heterogeneous data sources, including analytics data, web-scraped product content, and consumer clickstream panels, to monitor and predict product-level sales across global markets. As such, the integration of an automated categorization system was driven by the need to ensure taxonomy consistency across retailers and to support downstream analytics.

One of the major deployment challenges stemmed from the inherent heterogeneity of product categories across platforms. Retailers varied substantially in their taxonomy depth and granularity. Direct-to-consumer (DTC) brand sites and specialized fashion retailers often employed highly specific subcategories, while large generalist marketplaces tended to use broader, less detailed categories. Consequently, the scraped metadata used for categorization primarily derived from site breadcrumbs, displayed a wide range of inconsistent definitions. The same product could appear under divergent paths across domains, creating difficulties for cross-platform alignment. This lack of standardization impeded downstream analytics, such as attributing category-specific sales effects across platforms, and ultimately degraded the accuracy of EURWEB’s sales forecasting models.

To mitigate these issues, EURWEB adopted a two-stage inference pipeline informed directly by the results of this research. In the first stage, a lightweight unimodal RoBERTa model processes all newly identified products on a weekly basis. This stage provides efficient baseline categorization, taking advantage of RoBERTa’s strong text-based performance with minimal resource requirements. Products for which this model yields prediction confidence below a predefined threshold are automatically escalated to the second stage. In this stage, a multimodal RoBERTa + ViT model is employed, deployed within GPU-accelerated Docker containers. By leveraging both textual and visual features, this stage ensures high-fidelity categorization for ambiguous or complex products where text alone may be insufficient.

This cascading architecture offers a pragmatic balance between accuracy and operational efficiency. The majority of products are processed using the computationally lean RoBERTa model, while only a minority of low-confidence cases are escalated to the multimodal classifier. This design minimizes inference costs while still ensuring robust categorization quality. The deployment illustrates how advanced multimodal research outcomes can be adapted into scalable, production-ready pipelines. EURWEB is actively extending this two-stage framework beyond the ``Apparel \& Accessories'' sector to other verticals, such as consumer electronics and groceries, highlighting both the scalability and adaptability of the proposed approach in broader industrial contexts.

\section{Discussion and Conclusion}
\label{sec:discussionConclusion}

This study tackled the industrial challenge of multimodal hierarchical product categorization and taxonomy refinement, with a focus on cross-platform robustness and deployment at scale. Using 271,700 products from 40 e-commerce platforms, we compared unimodal and multimodal strategies, introduced a self-supervised recategorization pipeline, and validated the approach through real-world integration into EURWEB’s commercial analytics platform.

Our experiments confirmed the advantages of multimodality: CLIP-based late fusion achieved top accuracy, while dynamic masking preserved taxonomy consistency by preventing invalid category paths. Just as importantly, results revealed a practical trade-off: complex fusion methods excel with diverse training data, whereas simpler strategies generalize better to unseen platforms. For deployment, this indicates that industrial systems must not only optimize for peak benchmark accuracy but also for robustness under heterogeneous and resource-limited conditions. The proposed self-supervised recategorization pipeline further addressed structural gaps in existing taxonomies, generating fine-grained subcategories (e.g., within \textit{Shoes}) with strong purity levels. By reducing dependence on manual relabeling, this method provides a scalable path for platforms to evolve their taxonomies in response to market trends, aligning classification systems with dynamic product landscapes.

The EURWEB deployment demonstrates how research contributions translate into production. A two-stage pipeline: lightweight RoBERTa for the majority of cases and multimodal RoBERTa+ViT for low-confidence instances, balanced inference cost with classification fidelity. This cascading design reduced GPU overhead while improving category standardization, enabling more reliable forecasting and cross-platform analytics. Such architectures show how advanced multimodal models can be distilled into efficient, production-ready solutions.

Beyond immediate deployment, the framework has broader implications for e-commerce platforms of all sizes. Enhanced multimodal hierarchical classification improves product discoverability, search relevance, and personalized recommendations, while granular taxonomies strengthen inventory analytics, marketing optimization, and long-tail product discoverability. Smaller retailers, which often lack resources to build bespoke systems, can benefit from cross-platform models that transfer knowledge distilled from larger platforms. The semi-automated recategorization pipeline further provides a scalable mechanism for adaptive taxonomy maintenance, reducing manual costs and enabling taxonomies to evolve with emerging product trends. Looking forward, promising directions include extending these methods to other verticals such as electronics and groceries, integrating multilingual capabilities for global deployments, and combining SSL-driven recategorization with large language models (LLMs) for taxonomy generation and maintenance \cite{cheng2024commerce,chen-etal-2025-leveraging,gholamian-etal-2024-llm}. Additional work on extreme class imbalance, refined dynamic masking for multimodal architectures, and clustering methods that incorporate domain constraints will further strengthen robustness and industrial applicability.



%
%
%
\printbibliography
%







\end{document}